\pdfoutput=1

\documentclass[11pt]{article}

\usepackage[final]{acl}

\usepackage{times}
\usepackage{latexsym}

\usepackage[T1]{fontenc}

\usepackage[utf8]{inputenc}

\usepackage{microtype}

\usepackage{inconsolata}

\usepackage{graphicx}

\usepackage{amssymb}
\usepackage{textcomp}
\usepackage{hyperref}
\usepackage{url}
\usepackage{tipa}
\usepackage{tabularx}
\usepackage{subcaption}
\usepackage{booktabs}
\usepackage{multirow}
\usepackage{array}
\usepackage{setspace}
\title{Building a Functional Machine Translation Corpus for Kpelle}

\author{
  \textbf{Kweku Andoh Yamoah}\textsuperscript{1} \quad
  \textbf{Jackson Weako}\textsuperscript{2} \quad
  \textbf{Emmanuel J. Dorley}\textsuperscript{1} \\
  \textsuperscript{1}University of Florida \\
  \textsuperscript{2}Liberian Language Institute \\
  \texttt{kyamoah@ufl.edu}, \texttt{weakojackson@gmail.com}, \texttt{edorley@ufl.edu}
}

\begin{document}
\maketitle
\begin{abstract}
In this paper, we introduce the first publicly available English-Kpelle dataset for machine translation, comprising over 2,000 sentence pairs drawn from everyday communication, religious texts, and educational materials. By fine-tuning Meta’s No Language Left Behind (NLLB) model on two versions of the dataset, we
achieved BLEU scores of up to 30 in the Kpelle-to-English direction, demonstrating the benefits of data augmentation. Our findings align with NLLB-200 benchmarks on other African languages, underscoring Kpelle’s potential for competitive performance despite its low-resource status. Beyond machine translation, this dataset enables broader NLP tasks, including speech recognition and language modeling. We conclude with a roadmap for future dataset expansion, emphasizing orthographic consistency, community-driven validation, and interdisciplinary collaboration to advance inclusive language technology development for Kpelle and other low-resourced Mande languages.
\end{abstract}

\section{Introduction}

Several notable initiatives have sought to address the challenges of low-resource languages, particularly in Africa. Collaborative projects like Masakhane \citep{nekoto-etal-2020-participatory, orife2020masakhanemachinetranslation} have created and publicly released several machine translation datasets and baseline models for African languages \citep{nekoto-etal-2020-participatory, orife2020masakhanemachinetranslation, Nakatumba2024}. The Lacuna Fund has also played a vital role in accelerating the creation of openly accessible text and speech datasets for various African languages \citep{Nakatumba2024, Owusu2022Financial, bayelemabagamldataset2022, asmelash_teka_hadgu_machine_2022, wanjawa_kencorpus:_2024, adelani_few_2022}. Additionally, there is Meta's  "No Language Left Behind" (NLLB) project aimed to develop high-quality machine translation systems for over 200 languages, including many low-resource languages in Africa \citep{nllbteam2022languageleftbehindscaling}. Despite these efforts, languages such as the Kpelle language have not been explored, leaving the language marginalized in natural language processing (NLP) research. 

Kpelle is a language primarily spoken in Liberia and Guinea, with over one million speakers across these two countries \citep{MandeLanguages}. It is classified as a macro-language due to distinct variants—Liberian Kpelle and Guinean Kpelle—that, while closely related, constitute separate linguistic entities \citep{MandeLanguages}. Belonging to the Southwestern subgroup of the broader Mande language family, Kpelle is part of a larger linguistic family that includes approximately 70 languages spoken by at least 25 million native speakers and an additional 30 million second-language speakers throughout West Africa \citep{Konoshenko2008TonalSI, MandeLanguages}. Within Liberia specifically, Kpelle represents the largest indigenous language, spoken by approximately 20\% of the population \citep{MandeLanguages}.

Although Kpelle boasts a considerable number of speakers, it remains largely absent from digital platforms, including AI tools. Kpelle is a low-resourced language, which means the language lacks sufficient digital resources to support the development of NLP applications. Therefore, by extension, Kpelle faces the same challenges that are unique to low-resourced languages. These challenges include data scarcity \citep{kusampudi-etal-2021-corpus, maillard-etal-2023-small, Nakatumba2024, nguyen-etal-2022-kc4mt}, data quality(data limited to specific domains like religious texts) \citep{Nakatumba2024, maillard-etal-2023-small, kusampudi-etal-2021-corpus, nllbteam2022languageleftbehindscaling}, multilingualism, and dialectical variations(difficulty determining boundaries within dialects) \citep{konoshenko2024}.

To address this significant gap, we present the first-ever dataset for Kpelle. This dataset is designed for machine translation and language learning of Kpelle and English and vice versa. Our work aims to lay the foundations for intensive research for Kpelle and other low resource Liberian languages, enabling the development of NLP applications and solutions that can enhance the way speakers of the language interact with everyday technologies. This paper begins with an introduction highlighting our work's foundations and motivations. The continuing sections present the related work for machine translation for African languages.
We then present the history of the Kpelle language, examining its unique linguistic features. Following that, we discuss the dataset creation process and the corpus benchmarking using the NLLB model and the results obtained. Our contributions are as follows:\footnote{Dataset is made available at \url{https://huggingface.co/datasets/IARG-UF/English-Kpelle-Corpus}}
\textit{(a) Created a bilingual English-Kpelle corpus that has 3234 translation pairs. (b) The methodological data collection, cleaning, and alignment approach offers a replicable framework for other researchers working with low-resource languages. (c)  Benchmarked the dataset on NLLB achieving a BLEU of $\approx$ 30 for kpe\_Latn $\rightarrow$ eng\_Latn translation and a BLEU of $\approx$ 24 eng\_Latn $\rightarrow$ kpe\_Latn translation.}

\section{Related Work}
    \subsection{Review of Efforts in Low-Resource Language Datasets}
    
    The development of robust NLP tools for low-resource languages is limited by data scarcity, creating significant challenges for tasks like machine translation. Addressing this challenge has prompted various initiatives to expand language coverage and improve translation quality. Community-led projects like Masakhane have played a pivotal role in building datasets and models for African languages through a collaborative approach involving researchers and native speakers \citep{Nakatumba2024, akinfaderin-2020-hausamt}. The Lacuna Fund has further supported these efforts by funding the creation of open-source text and speech resources for African languages \citep{akinfaderin-2020-hausamt, Nakatumba2024, Owusu2022Financial, bayelemabagamldataset2022, asmelash_teka_hadgu_machine_2022, wanjawa_kencorpus:_2024, adelani_few_2022}. Meta’s ambitious “No Language Left Behind” (NLLB) project has made significant progress in building machine translation systems for over 200 languages, including many that are under-resourced \citep{nllbteam2022languageleftbehindscaling}. 	The NLLB \citet{nllbteam2022languageleftbehindscaling} used data mining to transform vast monolingual datasets into new training data for low-resource languages and employed new modeling approaches, like the Sparsely Gated Mixture of Experts, to improve translation quality \citep{nllbteam2022languageleftbehindscaling}. However, NLLB \citep{nllbteam2022languageleftbehindscaling}, like many other initiatives, primarily focuses on languages with established written standards, leaving languages with limited or no written traditions largely unaddressed .
    
    Beyond large-scale projects, creating specialized corpora has proven vital in addressing the data diversity and domain adaptation needs of specific languages and regions \citep{agyei_2024_low, maillard-etal-2023-small}. The Twi-2-ENG corpus from \cite{agyei_2024_low} is a recent example, providing a comprehensive resource for the Twi language, encompassing a wide range of genres relevant to Ghanaian Twi-speaking communities. This corpus aims to support NLP applications like machine translation and linguistic research by offering a searchable platform for accurate translations and a deeper understanding of Twi linguistics \cite{agyei_2024_low, george2024tangalenlp, williams-etal-2018-broad}. Another example is the LORELEI program, initiated by DARPA, which targets research and development of language technologies that aim to reduce the dependency on manually transcribed and translated corpora \citep{nguyen-etal-2022-kc4mt, agyei_2024_low, goyal2021flores101evaluationbenchmarklowresource}. This program has facilitated the collection of language samples and data for several African languages, including Hausa, Zulu, Yoruba, Twi, Somali, Swahili, and Wolof, contributing to the growth of language resources for these languages \cite{agyei_2024_low, goyal2021flores101evaluationbenchmarklowresource, nllbteam2022languageleftbehindscaling}.

    \subsection{Prior Work on the Mande Language Family}
    Existing NLP research on the Mande languages primarily focuses on individual languages, with limited cross-linguistic studies or comprehensive datasets representing the broader family \citep{MandeLanguages}. A few studies have investigated specific linguistic phenomena, such as the origin of the S-O-V-X word order \citep{MandeLanguages}, motion events in Bambara \citep{MandeLanguages}, and the evolution of tonal systems \citep{Konoshenko2008TonalSI, MandeLanguages}. Efforts in language documentation and corpus creation for Mande languages have also been undertaken \citep{george2024tangalenlp, Nakatumba2024,akinfaderin-2020-hausamt, nllbteam2022languageleftbehindscaling}. For instance, a grammatical sketch of Beng, a Southern Mande language, has been developed \citep{paperno_2014_sample}. Additionally, research on the Kakabe language, a Western Mande language, has focused on prosody in grammar \citep{vydrina:tel-03203594}. However, these efforts typically focus on individual languages or specific linguistic phenomena, and thus do not provide comprehensive resources or datasets necessary for cross-lingual NLP applications across the broader Mande language family.

    \subsection{Gap Filled by the Kpelle Dataset}
    The Kpelle Dataset aims to address a critical gap in the current research by providing the first, publicly available bilingual dataset for the Kpelle language. Despite being one of the most widely spoken languages in Liberia and Guinea, Kpelle remains severely underrepresented in NLP research, lacking any existing publicly available datasets. This absence stems from several factors, including Kpelle's status as a low-resource language with limited digital presence, the complexities arising from its dialectal variations across Guinea and Liberia \citep{Konoshenko2008TonalSI, MandeLanguages}, and the lack of standardized orthography \citep{konoshenko2024}. The dataset from this work will provide a much-needed resource for developing and evaluating NLP tools for Kpelle, enabling advancements in tasks like machine translation, language modeling, and speech recognition. By making this dataset publicly available, the project contributes to the broader goal of promoting language diversity and inclusion for African Languages.

\section{Overview of Kpelle}
\begin{figure*}[htbp]
        \centering
        \includegraphics[width=\linewidth, height=15em]{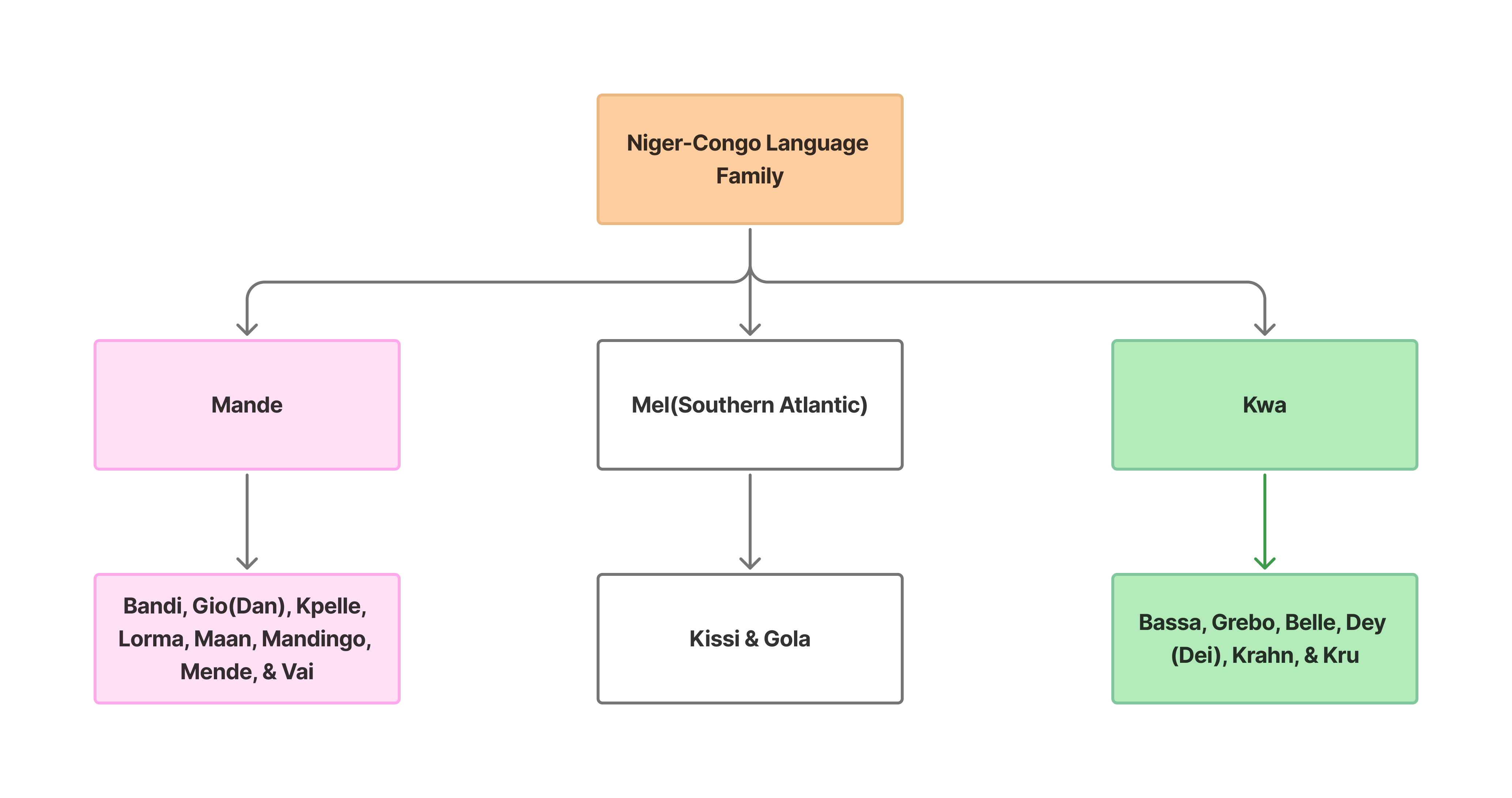}
        \caption{Overview of Liberian language family under the Niger-Congo Branch.}
        \label{fig:language_tree}
    \end{figure*}

    As previously mentioned, Kpelle belongs to the Southwestern Mande branch of the larger Mande language family. Figure \ref{fig:language_tree} illustrates how Kpelle fits within this broader linguistic context, demonstrating its relationship to other languages spoken throughout Liberia.

    Kpelle boasts of a rich oral tradition, with storytelling, proverbs, and songs playing a pivotal role in preserving the history and cultural values of the people \citep{Thach1981ALD}. Oral tradition has been key in maintaining the language across generations, especially since written text is limited \citep{Thach1981ALD}. Also, Kpelle faces challenges in representation and expansive linguistic research due to its low-resources status.
    
    Further, external influences have impacted the Kpelle language. In the 19th and 20th centuries, interactions with European colonizers and neighboring ethnic groups introduced new vocabulary into the language \citep{Thach1981ALD}. However, Kpelle has kept its core linguistic structure and continues to thrive as a means of communication and cultural identity for its speakers \citep{Thach1981ALD}.

\subsection{Linguistic Features}
    In this paper, we focus on \textbf{Liberian Kpelle} which exhibits distinct linguistic features that set it apart within the Mande Language family.
    
    \subsubsection{Phonetics}
    
    Kpelle uses a sound system with a rich array of consonants and vowels \citep{Thach1981ALD, MandeLanguages,konoshenko2024, thach1981kpelle}. Notably, it includes labiovelar stops such as \textipa{/gb/} and \textipa{/kp/}, which are said simultaneously at the velar and bilabial places of articulation and represent single consonant sounds \citep{Thach1981ALD, thach1981kpelle, MandeLanguages}. These sounds are relatively rare in global languages and contribute to Kpelle's unique phonological profile. 
    The vowel system in Kpelle has seven oral vowels and their nasal counterparts, making for a complex vocalic inventory \citep{MandeLanguages}. Dialectical variations influence pronunciation, particularly with the \textipa{/s/} sound \citep{Thach1981ALD}. In some regions, the \textipa{/s/} can resemble the English \textipa{/s/}; in others, it may sound like \textipa{/S/} (as in "ship") or \textipa{/h/} \citep{Thach1981ALD}. These forms of variations can pose difficulties for language learners.

    \subsubsection{Syntax}
    Kpelle follows a Subject-Verb-Object(SVO) sentence structure, which aligns with the syntactic patterns of many languages in the world, including English \citep{Thach1981ALD, MandeLanguages, Konoshenko2008TonalSI}. This syntactic structure facilitates the translation of Kpelle to English to some extent. Kpelle also distinguishes between dependent and independent nouns, akin to the idea of inalienable and alienable possession seen in other languages \citep{Thach1981ALD, MandeLanguages}. For example, body parts and kinship terms are treated differently grammatically compared to other nouns, affecting possessive constructions \citep{Thach1981ALD, MandeLanguages}.
    
    Modifiers in Kpelle usually follow the nouns they describe \citep{Thach1981ALD}, and the language employs postpositions rather than prepositions \citep{MandeLanguages}. Verb serialization is also a feature in Kpelle \citep{heine1984}, where multiple verbs are used sequentially to convey complex actions or events without conjunctions.

    \subsubsection{Grammar}
    Kpelle grammar has a complex system of pronouns that reflect distinctions in person, number, and sometimes gender \citep{Thach1981ALD, MandeLanguages}. The verb system marks tenses, aspect, and mood through affixes and particles \citep{Thach1981ALD, thach1981kpelle}. For example, there are specific markers for past, present, and future tenses and for completed and ongoing actions \citep{thach1981kpelle}.
    
    Noun classes in Kpelle are less prominent than in some other African languages but do exist and can affect agreement within the sentence \citep{MandeLanguages}. Kpelle employs emphatic particles like \textit{\textipa{"bé"} }to convey emphasis or focus within a sentence \citep{Thach1981ALD}. Since tone and stress are primarily used to convey lexical and grammatical meaning \citep{Thach1981ALD, thach1981kpelle, MandeLanguages}- these particles play an important role in adding nuance and emphasis without altering the tonal structure.

    \subsubsection{Tonality}
    Liberian Kpelle is a tonal language, meaning that the pitch at which a syllable is said can change the word's meaning entirely \citep{Thach1981ALD, thach1981kpelle, MandeLanguages, konoshenko2024, Konoshenko2008TonalSI}. Kpelle features three tone levels: high, mid, and low \cite{Thach1981ALD, MandeLanguages, Konoshenko2008TonalSI}. Tones can be level (staying the same throughout the syllable) or contour (changing pitch within the syllable) \citep{Thach1981ALD, thach1981kpelle, Konoshenko2008TonalSI}. This tonal system is essential for distinguishing words that are otherwise identical phonetically \citep{Konoshenko2008TonalSI}. For example, \citep{Konoshenko2008TonalSI} presents that "simple words in Kpelle form several groupings according to the tonal patterns which are assigned to these words lexically," and the groupings can be binned into categories known as \emph{tonal classes} \citep{Konoshenko2008TonalSI}. Also,  a single syllable pronounced with a high tone might mean one thing (\emph{lá, meaning mouth}); in a mid-tone, that same syllable communicates (\emph{la, meaning it}), while the same syllable with a low tone means something entirely different (\emph{là, meaning if}) \citep{Thach1981ALD}.

    Tone also plays a grammatical role in Kpelle, affecting verb tenses and aspects \citep{Konoshenko2008TonalSI}. Tonal patterns indicate whether an action is completed, ongoing, or habitual. This reliance on tone adds a layer of complexity to Kpelle learning and computational processing since accurate tonal representation is critical, especially for this work. Table \ref{tab:kpe-tones} presents the tones seen in Kpelle with examples.

\begin{singlespace}
    \begin{table}[!htpb]
        \centering
        \small
        \caption{Tonal Levels in Kpelle adapted from \cite{Thach1981ALD, a2024_libtralo}}
        \label{tab:kpe-tones}
        \begin{tabularx}{\linewidth}{@{}l c >{\centering\arraybackslash}X >{\centering\arraybackslash}X@{}}
            \toprule
            \textbf{Tonal Level} & \textbf{Mark} & \textbf{Kpelle Example} & \textbf{English Version} \\ 
            \midrule
            High            & \'\   & zóo      & native doctor \\
            Mid             & no mark/\=\   & tuna     & rain \\
            Low             & \`\   & ny\`{\textopeno}\textopeno   & be afraid \\
            High-Low        & \^\   & sâa      & today \\
            Mid-High-Low    & \^\   & tisô     & sneeze \\
            Low-High        & \v\   & k\v{\textopeno}   & to plant \\
            Nasal           & \~\   & s\~{a}   & to dance \\
            \bottomrule
        \end{tabularx}
    \end{table}
\end{singlespace}

    \subsubsection{Writing System}
    Historically, Kpelle has been primarily an oral language, but people have worked to develop writing systems that promote literacy and documentation. An example is the Kpelle syllabary created by Chief Gbili in the 1930s, an indigenous script designed to represent the sounds of Kpelle \citep{african_671_kpelle_2019}. However, few people use this script today \citep{african_671_kpelle_2019}.
    
    More commonly, Kpelle is written using Latin-based orthography \citep{MandeLanguages}. This system has been influenced by various scholars and linguists, such as William E. Welmers, who worked on developing practical orthographies for African languages in the mid-20th century \citep{Konoshenko2008TonalSI}. The Latin-based orthography often has diacritical marks to show tonal variations \citep{konoshenko2024, Thach1981ALD}; moreover, the lack of standardization leads to inconsistencies in written materials  \citep{konoshenko2024, Thach1981ALD}. 
    
    The Kpelle dictionary by \cite{Leideenfrost2005} incorporates tonal markings and provides valuable resources for language learners and researchers \cite{Thach1981ALD, Konoshenko2008TonalSI}. Materials from the Kpelle Literacy Center in Totota also use the Latin script to promote written literacy among native speakers of Kpelle \citep{Thach1981ALD}. The absence of a universally accepted orthography remains challenging, considering the variations between Liberian and Guinean Kpelle \cite{Thach1981ALD, Konoshenko2008TonalSI}.

\section{Dataset Creation}
Creating the English-Kpelle dataset involved planning and execution to ensure the data's relevance, accuracy, and cultural appropriateness. Our primary goal was to compile a corpus facilitating effective communication for individuals who may not speak Kpelle, particularly in everyday social interactions and essential services. This section outlines the data collection sources and methods, preprocessing steps, and translation alignment processes used in building the dataset.
\subsection{Data Collection}
\subsubsection{Sources}
The sources used in building the dataset covered a combination of practical and culturally relevant scenarios:

    \textbf{Travel and Tourism Phrases}. We identified common phrases and questions frequently asked by tourists and travelers when they visit a new location. Usually, due to their unfamiliar disposition to the place, we focused on phrases that covered greetings, inquiries about locations, costs, weather conditions, and other essential interactions. The phrases were sourced from the following respected travel and language teaching website: \emph{Business Insider's Travel Language Phrases}\citep{abadi_2018_ive}, \emph{EF Education First's Essential Phrases}\citep{b_2018_13}, \emph{Online Teachers UK's English for Tourism and Travel} \citep{writer_2017_english}, \emph{Go Overseas' Language Phrases Before Travelling} \citep{perez_2022_helpful}, \emph{Accessible Travel Phrasebook by Premiki} \citep{lonelyplanetgloballimited_2018_35}, and \emph{Wikivoyage's Afrikaans Phrasebook} \citep{towikivoyage_2005_west}.
    
    \textbf{Religious Texts}. Religious literature, like the Bible, often contains a wealth of translated material that can be valuable for language datasets. We added a few excerpts from publicly available religious texts that have been translated into Kpelle.

    \textbf{Educational Material}. Significant portions of the dataset were sourced from the book \emph{A Learner Directed Approach to Kpelle by Sharon V. Thach} \citep{Thach1981ALD}, \emph{English-Kpelle Dictionary, with a Grammar Sketch and English-Kpelle Finder List} \citep{Leideenfrost2005}, \emph{We Have Come To Learn Kpelle} \citep{ricks_2009_kwaa}. These resources had bilingual content, including matching English-Kpelle sentence pairs, standalone English paragraphs, and standalone Kpelle paragraphs.

\subsubsection{Methods}

    \textbf{Data Extraction}. We gathered a list of essential phrases and sentences relevant to everyday communication from the travel and tourism websites. These phrases were selected based on their frequency of use and utility in facilitating introductory interaction.

    \textbf{Translation}. For English or Kpelle paragraphs that did not have the corresponding translation, we engaged a native Kpelle speaker with linguistic expertise to provide accurate translations. 

    \textbf{Segmentation of Paragraphs}. In cases where the source material provided paragraphs rather than individual sentences, we segmented the text into sentence pairs. This approach increased the granularity of the dataset, making it suitable for machine translation tasks.

    \textbf{Expert Verification}. All translated sentences were reviewed by Kpelle language experts to verify the accuracy of the translations, the correctness of tone and grammar, and the appropriateness of context.

\subsection{Data Preprocessing}
\subsubsection{Cleaning}
The raw data collected contained inconsistencies such as typographical errors, informal language, and irrelevant content. We performed a thorough cleaning process to remove these anomalies. This included spell-checking, correcting grammatical errors, and eliminating duplicate entries. Special attention was given to resolving translation inconsistencies, especially where multiple translations existed for a single English phrase. The most accurate and contextually appropriate translation was selected based on expert advice.

\subsubsection{Normalization}
Given Kpelle's lack of a universally accepted writing system, we adopted the Latin-based orthography commonly used in educational materials and literacy programs. Diacritical marks were standardized to represent tonal variations accurately. All text data was encoded using UTF-8 Unicode to ensure compatibility across different platforms and tools. This was essential for preserving special characters and tonal markers unique to Kpelle. To maintain consistency, all text was converted to a standard case format, except where capitalization was necessary for proper nouns and the beginning of sentences.

\subsubsection{Segmentation}
The text was segmented into individual sentences using punctuation cues and linguistic rules specific to Kpelle. This process was manually verified due to the potential for misinterpretation by automated tokenizers not tailored to Kpelle. Within sentences, words were tokenized based on whitespace and morphological patterns. This facilitated subsequent processing tasks such as alignment and statistical analysis. Kpelle often uses contractions and compound words. These were carefully identified and treated according to linguistic guidelines to ensure accurate tokenization.

\section{Dataset Statistics and Analysis}
\subsection{Quantitative Overview}
The dataset has 3234\footnote{This count refers specifically to Version 2 of our dataset, which extends the initial 1,518 sentence pairs to 2,005 and increases word entries from 1,181 to 1,229.} entries corresponding to unique Kpelle-English translation pairs. Typically, each entry has one Kpelle sentence paired with its English equivalent; however, some entries contain sentences under a single translation unit (e.g., compound or complex sentences kept intact to preserve context). In total, the dataset contains 30,021 words (14,790 in Kpelle and 15,231 in English) and 4,369 sentences (2,202 in Kpelle and 2,167 in English). The longest sentences contain 70 Kpelle words and 49 English words, with the shortest being a single word in either language. Moreover, there are 4,702 unique Kpelle words and 3,579 unique English words, resulting in an overall vocabulary of 8,281 entries. These statistics make this the largest publicly available bilingual English–Kpelle resource to date.

\subsection{Sentence Length}
After our distribution analysis, we observed that most of the English sentences ranged from 3 to 15 words, with an average length of around 8 words per sentence. The Kpelle sentences vary more due to certain functional words' presence (or absence) and the possibility of encoding multiple concepts in a single phrase. However, the average Kpelle sentence length approximates 7 words, with most sentences falling between 3 and 12.

\begin{figure}[htbp]
  \centering
  \begin{subfigure}[b]{\columnwidth}
    \centering
    \includegraphics[width=\columnwidth,]{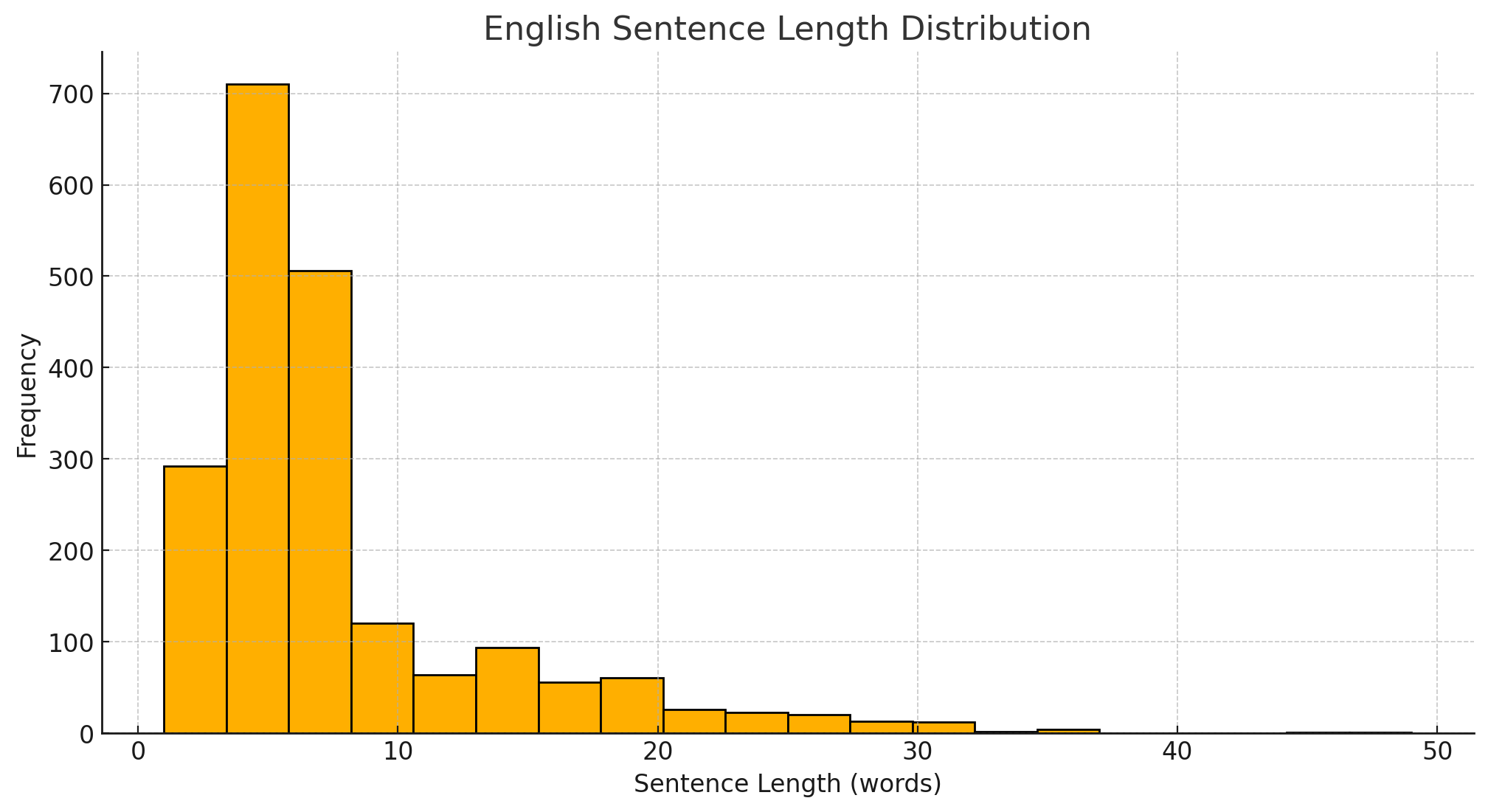}
    \caption{English sentence length distribution}
    \label{fig:eng_dist}
  \end{subfigure}
  \vspace{-1em} 
  \begin{subfigure}[b]{\columnwidth}
    \centering
    \includegraphics[width=\columnwidth, height=2.8cm]{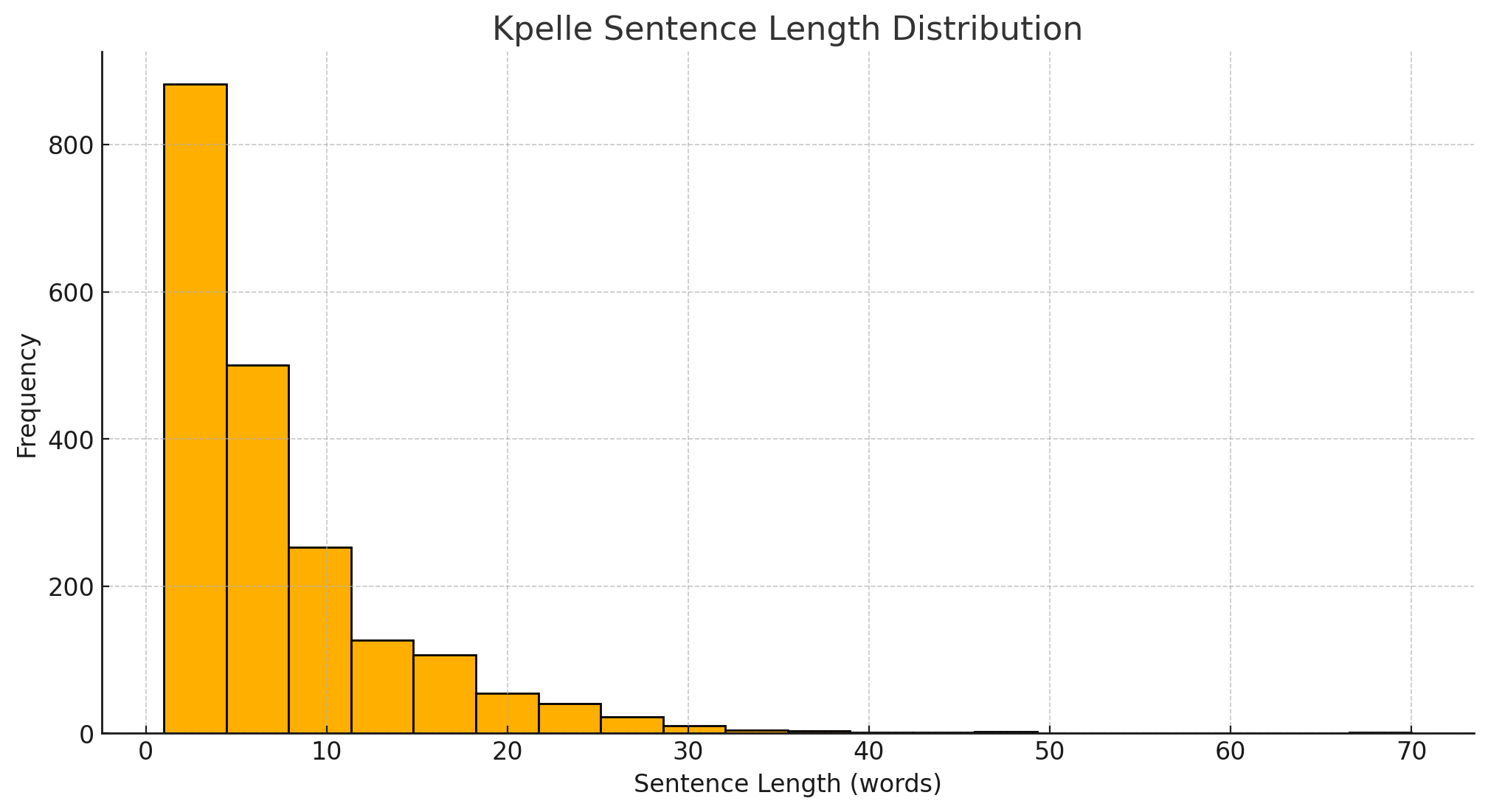}
    \caption{Kpelle sentence length distribution}
    \label{fig:kpelle_dist}
  \end{subfigure}
  \caption{Sentence length distributions for English (top) and Kpelle (bottom), illustrating the corpus’s inherent variability.}
  \label{fig:sentence_dists}
\end{figure}

The wide range of sentence lengths reflects the dataset’s inclusion of both simple and more complex utterances. Short, single-word sentences often correspond to exclamations, greetings, or short prompts, while longer sentences derive from religious or educational materials that contain embedded clauses and descriptive text.

\subsection{Vocabulary Frequency}

In terms of vocabulary, the top ten most frequent english words were \textit{man (116), good (93), town (93), want (83), go (75), going (65), one (61), house (59), baby (54), went (53)}. Similarly, the top ten most frequent Kpelle words were \emph{su (177), pâi (143), la (123), kaa (123), k\`{\textepsilon} (112), m{\textepsilon} (108)ni, p\^{\textopeno}ri (104), li (101), k{\textepsilon}  (99), k\^{\textepsilon}i (82)}. 

Even though we remove common stop words, frequent English words indicate a high presence of articles, pronouns, and commonly used verbs, mirroring everyday conversational usage. On the Kpelle side, repeated use of function words like \textit{\textbf{a, da},} and \textit{\textbf{e}} underscores similar syntactic necessities. These observations led to an English Hapax Legomena (words that appear once) of 1732 and a Kpelle Hapax Legomena of 2714.

A high number of hapax legomena suggests a rich and diverse vocabulary, but it also indicates that many words appear in the dataset with minimal frequency. This sparsity could pose challenges for certain NLP models, as low-frequency words often result in less robust embeddings and higher rates of out-of-vocabulary (OOV) tokens.

\subsection{Domain Coverage}
We conducted a keyword-based classification across common categories to understand the dataset's topical breadth. Table \ref{tab:sentence_distribution} shows that \textbf{Daily Conversation (664)} and \textbf{Household (214)} predominate, while underrepresented categories were \textbf{Religion (27)}, \textbf{Health (21)}, and \textbf{Education (14)}. It is worth noting that around 30\% of the dataset remains unclassified, reflecting idiomatic expressions and content not easily mapped to predefined categories. However, the broad coverage of the dataset, given the number of entries, ensures the dataset can serve a variety of use cases.

\begin{singlespace}
\begin{table}[htpb]
    \centering
    \small
    \begin{tabular}{lc}
        \toprule
        \textbf{Domain} & \textbf{Number of Sentences} \\
        \midrule
        Daily Conversation & 664 \\
        Household & 214 \\
        Business \& Finance & 142 \\
        Family & 93 \\
        Time \& Events & 91 \\
        Nature \& Environment & 80 \\
        General Purpose & 59 \\
        Religion & 27 \\
        Health & 21 \\
        Travel \& Tourism & 19\\       
        Education & 14 \\       
        Unclassified & 581\\
        \bottomrule
    \end{tabular}
    \caption{Distribution of Sentences by Domain}
    \label{tab:sentence_distribution}
\end{table}
\end{singlespace}

\subsection{Observations and Challenges}
Even though we adopt a standardized Latin-based script, Kpelle orthography's dynamic and evolving nature continues to introduce spelling and tone-marking variations throughout the dataset. These inconsistencies highlight the broader challenges of documenting a language with limited written traditions and underscore the importance of ongoing refinement in orthographic conventions. Additionally, the low representation of domains such as Religion, Health, and Education highlights future avenues for data collection to achieve more balanced coverage.  The distribution of topics also shows that key domains, such as Religion, Health, and Education, remain underrepresented, emphasizing potential areas for future data collection and corpus expansion to achieve more balanced coverage. 


\section{Experiments and Benchmarking}
\begin{figure*}[htbp]
  \centering
  \begin{subfigure}[b]{\columnwidth}
    \centering
    \includegraphics[width=\columnwidth]{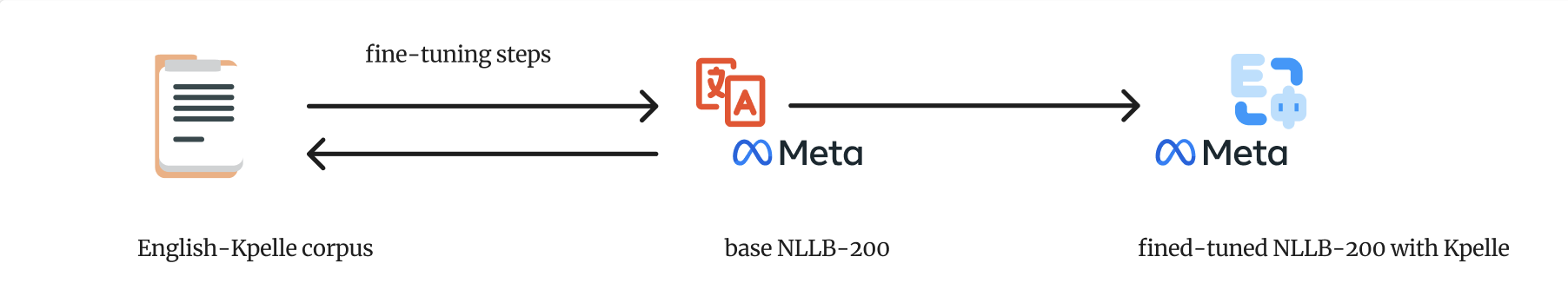}
    \caption{Fine-tuning NLLB-200 with Kpelle.}
    \label{fig:kpelle-finetuning}
  \end{subfigure}
  \vspace{-1em}
  \begin{subfigure}[b]{\columnwidth}
    \centering
    \includegraphics[width=\columnwidth]{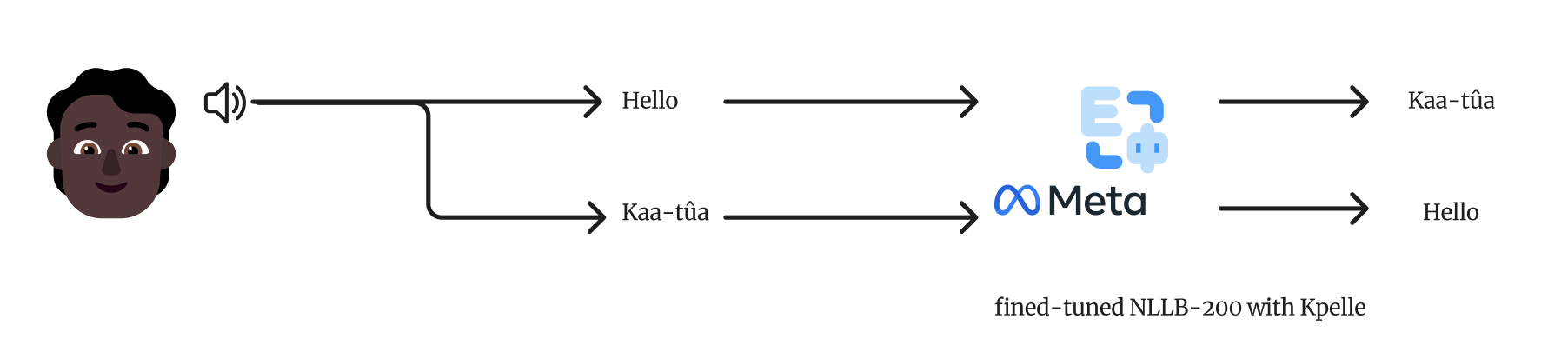}
    \caption{Translation example using the fine-tuned model.}
    \label{fig:machine-translation}
  \end{subfigure}
  \caption{NLLB-200 fine-tuning with Kpelle: (a) Model adaptation for bidirectional translation, and (b) a sample translation.}
  \label{fig:stacked}
\end{figure*}

This section presents our machine translation experiments and benchmarking using the NLLB model by \citep{nllbteam2022languageleftbehindscaling}. We describe our baseline models, outline the fine-tuning process, report quantitative results using standard evaluation metrics, and provide an analysis comparing our outcomes with previously reported NLLB-200 performance in other African languages. Figure \ref{fig:stacked} visually summarize this process.

\subsection{Baseline Models and Experimental Setup}
Given its strong performance across low-resource African languages, we leveraged Meta's NLLB model as a baseline. Our experiments focus on two Kpelle dataset versions:\textbf{Version 1 (V1)} contains 1,667 Kpelle and 1,638 English sentences (3,852 and 2,952 unique words). \textbf{Version 2 (V2}) benefits from data augmentation efforts, yielding 2,202 Kpelle and 2,167 English sentences (4,702 and 3,579 unique words). We aimed to assess how expanding the corpus (from 1,518 to 2,005 translation pairs) affects translation quality in both  English $\rightarrow$ Kpelle (eng\_Latn $\rightarrow$ kpe\_Latn) and Kpelle $\rightarrow$ English (kpe\_Latn $\rightarrow$ eng\_Latn). We split each dataset into two sets, train and test, according to a 9:1 ratio and hold out the test set. Then, these sets were fine-tuned for 10k, 30k, and 60k steps on top of NLLB using Adafactor as the optimizer with a batch size of 8, constrained by the memory requirements of the Quadro RTX 6000 GPU, training times ranged between 30 minutes to 6 hours, dependent on the number of steps and the version of the dataset. We trained a Kpelle-specific tokenizer (a SentencePiece model \cite{Kudo.2018}) on data from \citet{penedo2024fineweb-2} to handle out-of-vocabulary tokens and then enriched the standard NLLB tokenizer with any missing tokens, ensuring compatibility with the model's subword vocabulary. Finally, we used sacreBLEU \cite{post-2018-call} to measure BLEU, 1–4-gram precision, brevity penalty (BP), hypothesis/reference lengths, and chrF2++ to evaluate the fine-tuned model.

\subsection{Results}

\begin{singlespace}
\begin{table}[!htpb]
  \centering
  \small
  
   \resizebox{\columnwidth}{!}{%
      \begin{tabular}{@{}llccccccc@{}}
        \toprule
        \multirow{3}{*}{} & \multicolumn{8}{c}{\textbf{eng\_Latn $\rightarrow$ kpe\_Latn}} \\ 
        \cmidrule(lr){2-9}
        & \multirow{2}{*}{Steps} & \multirow{2}{*}{BLEU} & \multirow{2}{*}{chrF2++} & \multicolumn{4}{c}{Precision (1-4 grams)} & \multirow{2}{*}{BP} \\ 
        \cmidrule(lr){5-8}
        &  &  &  & 1-gram & 2-gram & 3-gram & 4-gram & \\
        \midrule
        \multirow{3}{*}{NLLB} 
          & 10k  & \textbf{24.09} & 38.24 & 49.3 & 28.7 & 20.5 & 16.8 & 0.913 \\
          & 30k  & \textbf{24.46} & 38.20 & 50.2 & 29.1 & 20.6 & 16.8 & 0.918 \\
          & 60k  & \textbf{24.00} & 38.19 & 50.1 & 28.2 & 19.5 & 16.1 & 0.930 \\
        \midrule
        \multirow{3}{*}{NLLB V2} 
          & 10k  & 19.80 & \textbf{38.26} & 49.6 & 25.4 & 15.5 & 10.0 & 0.942 \\
          & 30k  & 19.97 & \textbf{38.42} & 49.1 & 24.6 & 15.2 & 10.2 & 0.961 \\
          & 60k  & 20.79 & \textbf{38.83} & 51.5 & 26.9 & 16.9 & 11.4 & 0.915 \\
        \midrule
        \multirow{3}{*}{} & \multicolumn{8}{c}{\textbf{kpe\_Latn $\rightarrow$ eng\_Latn}} \\ 
        \cmidrule(lr){2-9}
        & \multirow{2}{*}{Steps} & \multirow{2}{*}{BLEU} & \multirow{2}{*}{chrF2++} & \multicolumn{4}{c}{Precision (1-4 grams)} & \multirow{2}{*}{BP} \\ 
        \cmidrule(lr){5-8}
        &  &  &  & 1-gram & 2-gram & 3-gram & 4-gram & \\
        \midrule
        \multirow{3}{*}{NLLB} 
          & 10k  & 23.16 & 38.29 & 42.5 & 24.7 & 18.6 & 14.7 & 1.000 \\
          & 30k  & 24.31 & 39.60 & 44.1 & 26.6 & 19.4 & 15.3 & 1.000 \\
          & 60k  & 23.65 & 39.41 & 43.1 & 25.2 & 18.9 & 15.2 & 1.000 \\
        \midrule
        \multirow{3}{*}{NLLB V2} 
          & 10k  & \textbf{26.39} & \textbf{40.22} & 50.0 & 30.6 & 20.9 & 15.2 & 0.999 \\
          & 30k  & \textbf{30.03} & \textbf{44.00} & 52.4 & 34.0 & 24.7 & 18.4 & 1.000 \\
          & 60k  & \textbf{30.28} & \textbf{44.28} & 53.4 & 34.5 & 24.8 & 18.3 & 1.000 \\
        \bottomrule
      \end{tabular}
    }
    \caption{NLLB performance when fine-tuned on two versions of the English-Kpelle dataset (V1 and V2) at 10k, 30k, and 60k steps. Metrics (BLEU, chrF2++, 1–4-gram precision, and BP) are reported for both eng\_Latn→kpe\_Latn and kpe\_Latn→eng\_Latn. Bold scores denote the best performance.}
    \label{tab:nllb_results}
\end{table}
\end{singlespace}

Table \ref{tab:nllb_results} summarizes the results for NLLB fine-tuned on V1 and V2 of the Kpelle dataset across 10k, 30k, and 60k training steps.

We observe that moving from \textbf{V1} (1,518 entries) to \textbf{V2} (2,005 entries) improved BLEU scores in some scenarios, particularly for kpe\_Latn $\rightarrow$ eng\_Latn translation at higher step counts (e.g., 30k, 60k). This outcome aligns with the broader expectation that additional in-domain data can boost model performance in low-resource settings. Further, we also observe that increasing the fine-tuning steps from 10k to 30k and 60k generally yielded incremental gains for both versions. However, the improvements were again more pronounced when translating from Kpelle to English. In contrast, eng\_Latn $\rightarrow$ kpe\_Latn translation showed modest gains, suggesting that further optimization may be necessary to achieve comparable results in translation quality for Kpelle.

\subsection{Analysis and Comparison with NLLB-200 Benchmarks}
Reports by \citet{nllbteam2022languageleftbehindscaling} highlight NLLB-200's performance across multiple African languages (e.g., Hausa, Igbo, Swahili, Yoruba).  As shown in Table \ref{tab:nllb_meta}, M2M-100, MMTAfrica, and NLLB-200 yield varying BLEU and chrF2++ scores for these languages. Given the differences in language structure, dataset sizes, and domain coverage, cross-lingual comparisons should be made cautiously. However,  \textbf{the scores we observe for Kpelle (BLEU in the range of 20–30 depending on the direction and training steps) are generally consistent with NLLB-200's range for other African languages.}.

\begin{table}[htbp]
    \centering
    \small
    \resizebox{\columnwidth}{!}{%
        \begin{tabular}{l@{\quad}ccc@{\quad}ccc}
            \toprule
             & \multicolumn{3}{c}{\textbf{eng\_Latn--xx}} 
             & \multicolumn{3}{c}{\textbf{xx--eng\_Latn}} \\
            \cmidrule(lr){2-4}\cmidrule(lr){5-7}
             & MMTAfrica & M2M-100* & NLLB-200 & MMTAfrica & M2M-100* & NLLB-200 \\
            \midrule
            \textbf{hau\_Latn} & -/- & 4.0/-   & \textbf{33.6/53.5} & -/- & 16.3/-  & \textbf{38.5/57.3} \\
            \textbf{ibo\_Latn} & 21.4/37.2 & 19.9/- & \textbf{25.8/41.4} & 15.4/38.9 & 12.0/-  & \textbf{35.5/54.4} \\
            \textbf{lug\_Latn} & -/- & 7.6/-   & \textbf{16.8/39.8} & -/- & 7.7/-   & \textbf{27.4/46.7} \\
            \textbf{luo\_Latn} & -/- & 13.7/-  & \textbf{18.0/38.5} & -/- & 11.8/-  & \textbf{24.5/43.7} \\
            \textbf{swh\_Latn} & 40.1/53.1 & 27.1/- & 37.9/\textbf{58.6} & 28.4/56.1 & 25.8/- & \textbf{48.1/66.1} \\
            \textbf{wol\_Latn} & -/- & 8.2/-   & \textbf{11.5/29.7} & -/- & 7.5/-   & \textbf{22.4/41.2} \\
            \textbf{xho\_Latn} & 27.1/44.9 & -/- & \textbf{29.5/48.6} & 21.7/48.6 & -/- & \textbf{41.9/59.9} \\
            \textbf{yor\_Latn} & 12.0/28.3 & 13.4/- & \textbf{13.8/25.5} & 9.0/30.6 & 9.3/- & \textbf{26.6/46.3} \\
            \textbf{zul\_Latn} & -/- & 19.2/-  & \textbf{36.3/53.3} & -/- & 19.2/-  & \textbf{43.4/61.5} \\
            \bottomrule
        \end{tabular}
    }
    \caption{BLEU/chrF2++ performance on selected African languages (eng\_Latn \(\leftrightarrow\) xx) for MMTAfrica, M2M-100*, and NLLB-200 from \citep{nllbteam2022languageleftbehindscaling}.}
    \label{tab:nllb_meta}
\end{table}

Our kpe\_Latn $\rightarrow$ eng\_Latn best BLEU of \textbf{30.28} at 60k steps surpasses NLLB-200's lower-bound performances (22.4 BLEU on Wolof), mid-range(24.5 BLEU on Luo, 26.6 BLEU on Youruba, 27.4 BLEU on Luganada) results, though it remains below the model's high performance (48.1 BLEU on Swahili). The eng\_Latn $\rightarrow$ kpe\_Latn translation lags slightly behind, reaching approximately \textbf{24.46} BLEU with V1 at 30k steps. This result is comparable and higher to NLLB-200's results ($\approx$ 25.8 BLEU in some languages) but lower than its highest observed values (37.9 BLEU in Swahili). Kpelle translations have the potential to reach NLLB-200's highest performance levels with further data augmentation and fine-tuning. However, language-specific nuances, such as Kpelle's orthographic variations, limited standardization, and relatively small corpus size, currently limit model performance.

\section{Conclusion}
This paper introduced the first publicly available English-Kpelle dataset for machine translation. Our corpus has over 2,000 translation pairs from diverse domains, such as daily conversation, household activities, and religious texts. We demonstrated the dataset's usability by fine-tuning Meta's NLLB model on two corpus versions. Our experiments revealed that data augmentation significantly benefits translation performance, particularly in the Kpelle-to-English direction at higher fine-tuning steps. These findings highlight the importance of domain-specific data expansion in enhancing translation quality for low-resource languages. Moreover, comparative analysis against reported NLLB-200 results highlights the potential for Kpelle NLP systems to achieve competitive performance levels, given continued data curation and iterative fine-tuning.

\section{Limitations}
\begin{enumerate}
    \item \textbf{Dataset Expansion and Domain Coverage:} 
    While we have made progress in building a representative English-Kpelle dataset, some gaps remain. Future efforts could focus on collecting domain-specific materials from underrepresented categories such as nature, environment, and specialized technical fields to enhance the domain coverage of the dataset further. Adding more varied dialectal data is also essential to capture the linguistic richness of Kpelle more comprehensively.

    \item \textbf{Broader NLP Applications:}
    Beyond machine translation, the dataset can be a foundation for other NLP tasks, including speech recognition, language modeling, and sentiment analysis. We intend to explore these avenues, building on the groundwork established here to develop robust and context-aware Kpelle language tools.

    \item \textbf{Limited Cross Model Evaluation:}
    Our current evaluation relies exclusively on fine-tuning Meta’s NLLB model. While NLLB provides a strong baseline for low-resource translation, this restricts our understanding of how the dataset performs across diverse architectures. As future work, we plan to benchmark an expanded version of the dataset on additional models, including M2M-100 and BLOOMZ, to better assess transferability and generalization. We also intend to incorporate complementary evaluation metrics, such as TER and METEOR, to provide a more comprehensive analysis of model performance.

    \item \textbf{Lack of Qualitative Error Analysis:}
    The current scope of this work sought to present the first English-Kpelle dataset and understand Kpelle's potential by benchmarking on a strong baseline like Meta's NLLB model. Given this, we failed to conduct a qualitative error analysis on the translation generated for the held-out test set. In future work, we plan to introduce human evaluation loops where native Kpelle speakers assess translation quality and identify systematic errors. This feedback will guide targeted model improvements and support a more fine-grained understanding of the dataset’s linguistic challenges.
    
\end{enumerate}

\subsection{Call to Action}
We invite researchers, linguists, and language technology enthusiasts to collaborate in expanding and refining this dataset. By contributing additional Kpelle text resources, validating translations, or developing novel NLP techniques, the research community can help bridge the digital divide faced by low-resource languages. We hope the work presented here will spark renewed interest in Kpelle and other underrepresented Mande languages, ultimately driving innovation and inclusivity in multilingual NLP.

\section{Acknowledgments}
We acknowledge the contributions of 
Mr. Aaron D. Y. Pope, Cuttington University, and Mr. Better Jallah,  University of Liberia, who served us our Kpelle translation experts, providing Kpelle translation pairs for gathered English sentences and words.

\bibliography{custom}




\end{document}